\documentclass[runningheads]{llncs}

\usepackage{eccv}

\usepackage{eccvabbrv}

\usepackage{graphicx}
\usepackage{booktabs}
\usepackage{caption}

\usepackage[accsupp]{axessibility}  

\usepackage{hyperref}

\usepackage{orcidlink}

\begin{document}

\title{ASTAD: Asymmetric Style Transfer for Synthetic-to-Real Adaptation in Autonomous Driving} 

\titlerunning{ASTAD}

\author{Dingyi Yao\orcidlink{0009-0008-5185-1282} \and
Xinqi Zhang\orcidlink{0009-0009-7902-6863} \and
Lihui Peng\orcidlink{0000-0001-7363-6374}\thanks{Corresponding author.} \and
Jianming Hu\orcidlink{0000-0001-8065-7309} \and
Danya Yao\orcidlink{0000-0001-5032-6322} \and
Yi Zhang\orcidlink{0000-0001-5526-866X}
}

\authorrunning{D.Yao et al.}

\institute{ Department of Automation, Tsinghua University, Beijing 100084, China
\email{\{ydy24,zhang-xq24\}@mails.tsinghua.edu.cn}; \email{\{lihuipeng, hujm, yaody, zhyi\}@tsinghua.edu.cn}}

\maketitle

\begin{abstract}
Synthetic data mitigates the data scarcity problem in autonomous driving perception. However, the synthetic-to-real gap leads to performance degradation, hindering real-world model generalization. Although current methods leverage diffusion models for photorealistic style transfer to bridge this gap, they critically ignore a practical asymmetry: while synthetic data possesses perfect pixel-level annotations, real-world style reference images generally lack corresponding labels. Consequently, existing methods relying on symmetric semantic guidance suffer from either prohibitive annotation costs or severe semantic misalignment. To address this dilemma, we formally propose a novel task: Asymmetric Style Transfer for Autonomous Driving (ASTAD), which requires semantically consistent transfer using only labeled synthetic content and unlabeled real-world references. We further introduce the ASTModel, a training-free two-stage framework designed to bridge this domain gap under asymmetric constraints. ASTModel first extracts a coarse semantic prior from the unlabeled target, followed by dynamic prior refinement and class-consistent style injection during the denoising process. Extensive experiments demonstrate that ASTModel significantly outperforms existing methods in downstream perception utility and structural fidelity, while offering a 3.2$\times$ inference speedup. This work aligns synthetic-to-real adaptation with practical constraints, holding the potential to accelerate the scalable deployment of robust autonomous driving systems. Code: \url{https://github.com/Dingyi-Yao/ASTAD}.
\end{abstract}

\section{Introduction}
\label{sec:intro}

Deep learning-based perception systems constitute the backbone of modern autonomous driving, enabling vehicles to interpret complex environments with high precision. However, the success of these systems is predicated on the availability of large-scale, high-density, and fine-grained semantic annotations. Acquiring high-quality annotations for real-world datasets is prohibitively time-consuming. For example, the annotation of a single Cityscapes\cite{cityscapes} image requires roughly 1.5 hours\cite{2}. Consequently, synthetic data generated by simulators has emerged as a complementary solution. It enables the rapid and cost-effective generation of large-scale datasets while providing error-free annotations\cite{3}. For instance, the GTA dataset\cite{gta} leverages the interception of communication between the game engine and graphics hardware to automatically generate precise pixel-level semantic labels. 

\begin{figure}[t]
  \centering
 \includegraphics[width=0.75\linewidth]{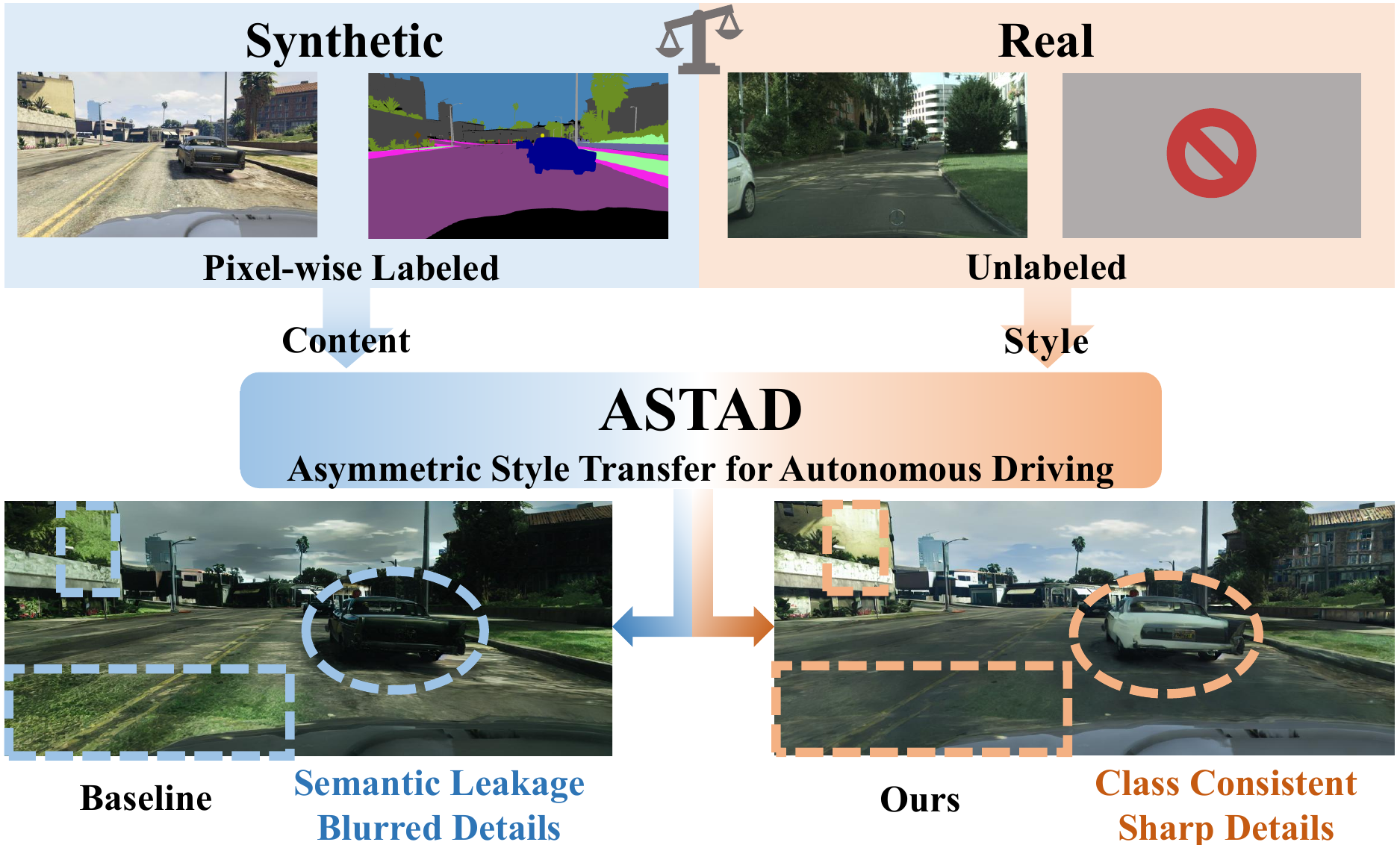}
   \caption{Illustration of ASTAD and the motivation of ASTModel.
  \textbf{Top:} ASTAD targets asymmetric synthetic-to-real style transfer, where synthetic content images have pixel-wise labels while real style images are unlabeled. \textbf{Bottom:} Without style-side semantic guidance, existing baselines suffer from semantic leakage and blurred scene details, whereas ASTModel enables class-consistent style injection, producing target-style images with preserved driving-scene geometry.}
   \label{fig:tu1}
\end{figure}

Despite these advantages, the practical deployment of synthetic data is bottlenecked by the inherent distribution shift between synthetic and real-world domains. This synthetic-to-real gap primarily stems from the limitations of simulation engines, including approximated physical rendering, idealized material textures\cite{5}. Consequently, models trained purely on synthetic distributions suffer from severe performance degradation when deployed in the real world\cite{6}. Bridging this critical gap is therefore essential to translate synthetic scalability into safe, real-world reliability.

To mitigate this, synthetic-to-real adaptation has become a crucial research direction. A promising direction is generating "target-style synthetic data", images that retain the precise layout of the simulation but exhibit the visual characteristics of the real world. In this context, Diffusion Models have emerged as a powerful paradigm\cite{diffusion,ldm}. Through an iterative denoising process, they demonstrate superior capabilities in accurately modeling complex data distributions and generating high-fidelity details\cite{tcvst4,tcvst5,tcvst6}. With their zero-shot generalization capabilities derived from large-scale pre-training, diffusion models offer a powerful mechanism to inject real-world styles into synthetic content\cite{crossimage, zstar,styleid,eyeforaneye, CACTIF}.

However, a critical contradiction remains overlooked in existing diffusion-based stylization methods: \textbf{the inherent information asymmetry between content and style}. Existing methods \cite{CACTIF} typically require explicit semantic segmentation maps for both content and style images to guide the transfer process and prevent semantic misalignment (\eg, placing vegetation textures on roads). Yet a practical dilemma arises: while pixel-perfect segmentation labels are effortlessly available for synthetic content images, obtaining corresponding dense annotations for real-world style reference images is difficult. 

We formally term this fundamental imbalance \textbf{"Asymmetry"}: synthetic content images are paired with pixel-level semantic masks, whereas real-world style images are available only as raw images without dense semantic labels. Because of this, existing methods often fail in practical autonomous driving scenarios, either by relying on requiring expensive manual annotation for style images or by suffering from severe cross-class pollution when such labels are absent, as shown in \cref{fig:tu1}.

To address this dilemma, we formally propose a novel task: \textbf{Asymmetric Style Transfer for Autonomous Driving (ASTAD)}. ASTAD aligns synthetic-to-real adaptation with practical constraints by requiring semantically consistent transfer using only labeled synthetic content and a raw, unlabeled real-world style image. To tackle this, we introduce \textbf{ASTModel}, a training-free, two-stage framework. Our main contributions are summarized as follows:
\begin{itemize}
  \item We formalize the ASTAD task, identifying the critical label asymmetry in synthetic-to-real adaptation and establishing a realistic problem setting where target style labels are unavailable.
  \item We introduce the ASTModel to resolve the asymmetry constraints. It incorporates Prototype-Guided Style Semantic Prior Extraction to address the semantic cold-start problem via foundation models, and integrates Multi-Layer Semantic Voting, Semantically Constrained Adaptive Attention Filtering and Pixel-Proportion Modulated Hybrid AdaIN, to robustly enforce semantic consistency and correct mismatches during the denoising process.
  \item Extensive experiments demonstrate that ASTModel generates images with enhanced semantic consistency, with its inference speed substantially improved.
\end{itemize}

\section{Related Work}
\subsection{General Neural Style Transfer}
Early Neural Style Transfer (NST) optimized feature Gram matrices \cite{rw1}, later accelerated by feed-forward networks utilizing Adaptive Instance Normalization (AdaIN) \cite{rw2}. To address AdaIN's loss of local details and structural "content leak", subsequent works introduced point-wise attention \cite{adaattn}, Transformer architectures \cite{stytr2,splicevit}, and reversible flows \cite{artflow}. Concurrently, GANs formulated stylization as image-to-image translation across paired \cite{rw4}, unpaired \cite{rw5,rw5-2}, multimodal \cite{rw6}, and multi-domain \cite{rw7-1,rw7-2} settings, or via explicit structure-texture latent swapping \cite{swapping}, though often bounded by training instability \cite{tcvst8}. Recently, Diffusion Models\cite{ldm} dominated generative tasks \cite{tcvst4,tcvst5,tcvst6}. Strategies adapting diffusion models for style transfer can generally be categorized into training-based and training-free methods. Training methods\cite{stylediffusion,blora,inst,vct} extract style concepts from reference images into conditional embeddings to guide diffusion-based style transfer. Conversely, training-free methods modify the denoising process directly. Early approaches guided reverse diffusion using disentangled ViT representations \cite{diffuseit}. Later methods swapped self-attention keys and values to inject style \cite{styleid, crossimage, zstar}, while advanced approaches further enforce semantic consistency by rearranging features based on dense correspondences \cite{eyeforaneye}.

Nevertheless, significant differences still exist between these generic artistic stylization methods and the specific demands of autonomous driving. First, artistic transfer typically focuses on object-centric images with a clear foreground-background separation. In contrast, driving scenes utilize high semantic density, characterized by multi-instance occurrences and diverse categories within a single view. Second, the primary metric for artistic transfer is perceptual aesthetics, often tolerating or even encouraging geometric deformations. Conversely, autonomous driving mandates the rigorous preservation of structural and semantic boundaries, as any distortion directly compromises the utility of the data for downstream perception tasks. Third, the autonomous driving field possesses a unique advantage: the inherent availability of segmentation maps, particularly for synthetic data. This facilitates stylization guided by explicit semantic information, a critical capability that general artistic methods typically overlook. These limitations necessitate specialized adaptation strategies tailored for autonomous driving perception tasks.

\subsection{Style-Based Synthetic-to-Real Adaptation}
To bridge the gap between general stylization and the rigorous demands of autonomous driving, researchers have developed specialized style-based adaptation methods. The primary goal is to mitigate the distribution shift between synthetic simulation and the real world by injecting realistic textures into annotated synthetic data. Pioneering works like CyCADA \cite{cycada} and SHADE \cite{shade} advanced synthetic-to-real adaptation by enforcing semantic consistency and enhancing style diversity. However, these methods often struggle with training instability or limited generative fidelity compared to modern standards. Diffusion Models have significantly elevated generation quality. Methods such as DGInStyle \cite{dginstyle}, SimGen \cite{simgen}, and Weather-Diff \cite{weatherdiff} achieve photorealism by integrating structural guidance (\eg, semantic masks, simulator layouts) and adapting to specific domains through fine-tuning. However, these training-based approaches are computationally intensive and inflexible, requiring a dedicated model training for each specific target domain. To bypass training costs, recent research employs training-free attention injection. Notably, CACTIF \cite{CACTIF} mitigates semantic inconsistencies by utilizing class-aware mechanisms, employing explicit segmentation masks to strictly confine style transfer to corresponding semantic regions. 

However, these methods rely on a "Symmetric Information" assumption, requiring dense annotations for both source content and target style. This is impractical for autonomous driving, where real-world style labels are often unavailable. In this asymmetric setting, symmetric methods fail to distinguish semantic regions, leading to severe cross-class semantic leakage. Addressing this unresolved dilemma motivates our proposed work.

\section{Methodology}
\subsection{Task Formulation}
\label{sec:taskform}

Let $\mathcal{D}_{src} = \{(I_c^i, M_c^i)\}_{i=1}^N$ denote the synthetic source domain, containing $N$ content images $I_c$ with pixel-wise semantic masks $M_c $. Conversely, the real-world target domain is represented by a style reference image $I_s$.

Existing diffusion-based style transfer methods, such as CACTIF\cite{CACTIF}, typically operate under a Symmetric Information Assumption. Formally, the symmetric transfer function $\Phi_{sym}$ is defined as $I_{out} = \Phi_{sym}(I_c, M_c, I_s, M_s)$.

To align with practical deployment constraints, we introduce the Asymmetric Style Transfer for Autonomous Driving (ASTAD) task. In this paper, ``asymmetric'' specifically refers to the annotation-level information imbalance between the labeled synthetic content image and the unlabeled real-world style image. That is, the synthetic source provides $(I_c, M_c)$, whereas the style image is only available as a raw image $I_s$, implying $M_s = \emptyset$.

Under this annotation-asymmetric setting, the objective of ASTAD is to generate a stylized image $I_{out}$ that inherits the visual characteristics of the real-world style image $I_s$ while preserving the semantic layout specified by the synthetic mask $M_c$. Formally, the asymmetric transfer function $\Phi_{asym}$ is defined as $I_{out} = \Phi_{asym}(I_c, M_c, I_s, \emptyset)$.

\subsection{Framework Overview}

\begin{figure*}[t]
  \centering
 \includegraphics[width=1.0\linewidth]{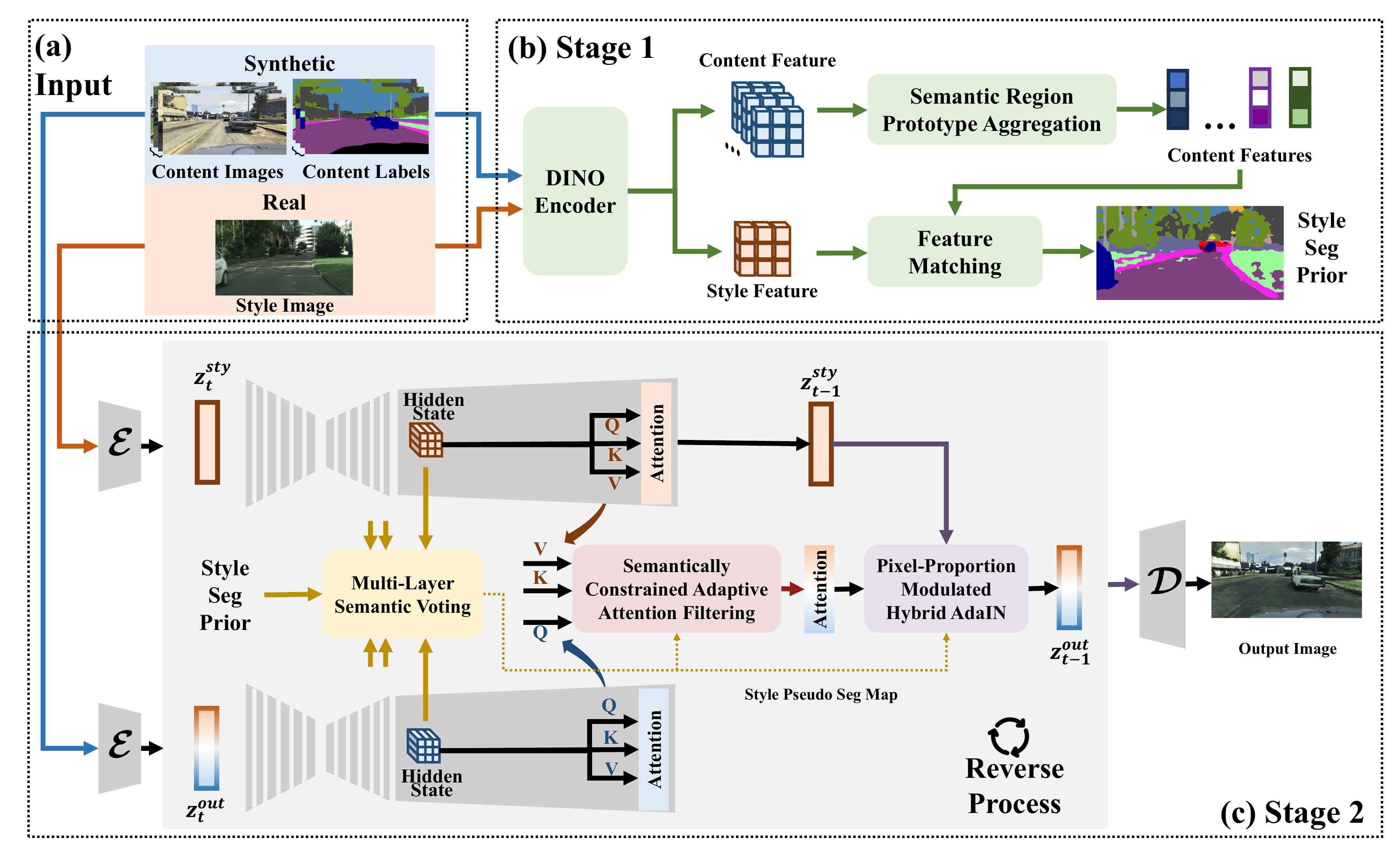}
   \caption{Overview of the proposed ASTModel. The pipeline operates as a training-free, two-stage framework to address asymmetric constraints. \textbf{(a) Inputs:} Abundant labeled synthetic data and a single unlabeled real-world style image. \textbf{(b) Stage I (Implicit Semantic Discovery):} Extracts a coarse Style Segmentation Prior by leveraging the semantic correspondence between synthetic prototypes and style features in the DINO space. \textbf{(c) Stage II (Asymmetric Style Injection):} Refines this prior via the Multi-Layer Semantic Voting during the diffusion reverse process, and performs style injection using Semantically Constrained Adaptive Attention Filtering and Pixel-Proportion Modulated Hybrid AdaIN.
  }
   \label{fig:pipeline}
\end{figure*}

As illustrated in \cref{fig:pipeline}, we propose ASTModel, a training-free framework for synthetic-to-real adaptation. Our pipeline operates in two stages: First, Stage I (\cref{subsec:stage1}) extracts a coarse prior from the unlabeled target via foundation model prototype matching. Next, Stage II (\cref{subsec:vote}--\cref{subsec:adain}) refines this prior during the diffusion reverse process using a Multi-Layer Semantic Voting (\cref{subsec:vote}). This dynamically sharpened guidance then strictly governs two novel injection modules: Semantically Constrained Adaptive Attention Filtering (\cref{subsec:filter}) to prevent semantic leakage, and Pixel-Proportion Modulated Hybrid AdaIN (\cref{subsec:adain}) to robustly align class-aware feature statistics.
\subsection{Prototype-Guided Style Semantic Prior Extraction}
\label{subsec:stage1}

To extract a coarse semantic prior for the unlabeled style image, we leverage the robust feature space of a frozen pre-trained Vision Transformer, DINO\cite{dinov2}, as our feature extractor $\mathcal{F}$.
Given a synthetic content image $I_c$ and a real-world style image $I_s$, we extract their $L_2$-normalized patch-level feature representations $\mathbf{F}_c = \mathcal{F}(I_c)$ and $\mathbf{F}_s = \mathcal{F}(I_s)$. 

We leverage the available synthetic annotations $M_c$ to construct robust semantic prototypes. For each semantic category $k \in \{0, \dots, C-1\}$, we define the semantic region $\mathcal{R}_k = \{(h, w) \mid M_c(h, w) = k\}$. To compute a class-specific prototype $\mathbf{p}_k$, we aggregate the feature vectors within each region
\begin{equation}
  \mathbf{p}_k = \text{Normalize}\left( \frac{1}{|\mathcal{R}_k|} \sum_{(h, w) \in \mathcal{R}_k} \mathbf{F}_c(h, w) \right)
  \label{eq:4}
\end{equation}

Treating these synthetic prototypes as semantic anchors, we formulate style parsing as a dense nearest-neighbor matching in the shared DINO space. The coarse semantic prior $M_{prior}$ is obtained by assigning each spatial location to the category with the highest cosine similarity
\begin{equation}
  M_{prior}(h, w) = \underset{k}{\arg\max} \,  \langle \mathbf{F}_s(h, w), \mathbf{p}_k \rangle
  \label{eq:6}
\end{equation}

While effective for global structural layout, $M_{prior}$ is inherently bounded by the patch-level resolution of the architecture, resulting in jagged boundaries. This coarse blueprint is thus forwarded to Stage II for dynamic refinement.

\subsection{Multi-Layer Semantic Voting}
\label{subsec:vote}
Stage II operates within the reverse process of a pre-trained Latent Diffusion Model (LDM) \cite{ldm}. While the Stage I prior $M_{prior}$ provides a semantic blueprint, it lacks spatial precision. Conversely, the high-resolution decoder layers of the U-Net\cite{unet} capture rich fine-grained structures. To synergize these complementary strengths, we introduce a conservative Multi-Layer Semantic Voting to dynamically refine the semantic guidance.

At each timestep $t$, we extract feature maps from a set of high-resolution decoder layers $\mathcal{L}$. To interpret these latent features semantically, we inherit the prototype-based paradigm established in Stage I. For a specific layer $l$, we compute layer-specific prototypes $p_k^l$ from the content features $f_{content}^l$ guided by $M_c$, reusing the aggregation logic defined in \cref{eq:4} and \cref{eq:6}. We then perform nearest-neighbor matching in this shared feature space to generate an instantaneous pseudo-segmentation map $\hat{M}_l^t$
\begin{equation}
  \hat{M}_l^t = \underset{k}{\arg\max} \langle \mathbf{f}_{style}^l, \mathbf{p}_{k}^l \rangle
\end{equation}
where $\mathbf{p}_{k}^l$ represents the prototype for class $k$ at layer $l$. 

Due to the stochastic nature of the diffusion process, a single layer's prediction $\hat{M}_l^t$ may be unstable. However, semantic structures that are intrinsic to the image should manifest consistently across different layers. We define a Consensus Mask $\mathcal{C}^t$, which activates only when all layers unanimously agree on the classification result, indicating high confidence

\begin{equation}
  \mathcal{C}^t(u, v) = \mathbb{I}\left[ \forall i, j \in \mathcal{L}, \hat{M}_i^t(u, v) = \hat{M}_j^t(u, v) \right]
\end{equation}

The refined Style Pseudo-Segmentation Map $\hat{M}_{s}^t$ is then synthesized by dynamically updating the static Stage I prior only where the diffusion features exhibit high-confidence consensus
\begin{equation}
  \hat{M}_{s}^t(u, v) = 
\begin{cases} 
\hat{M}_l^t(u, v) & \text{if } \mathcal{C}^t(u, v) = 1  \\
M_{prior}(u, v) & \text{otherwise}
\end{cases}
\label{eq:mask}
\end{equation}

This mechanism ensures that the system benefits from the fine-grained details captured by the diffusion model while maintaining the semantic stability provided by the foundation model. This refined map $\hat{M}_{s}^t$ serves as the ground truth for the subsequent filtering and alignment modules, ensuring that style injection is guided by sharp, dynamically corrected semantic boundaries.
\subsection{Semantically Constrained Adaptive Attention Filtering}
\label{subsec:filter}

\begin{figure}[t]
  \centering
 \includegraphics[width=0.6\linewidth]{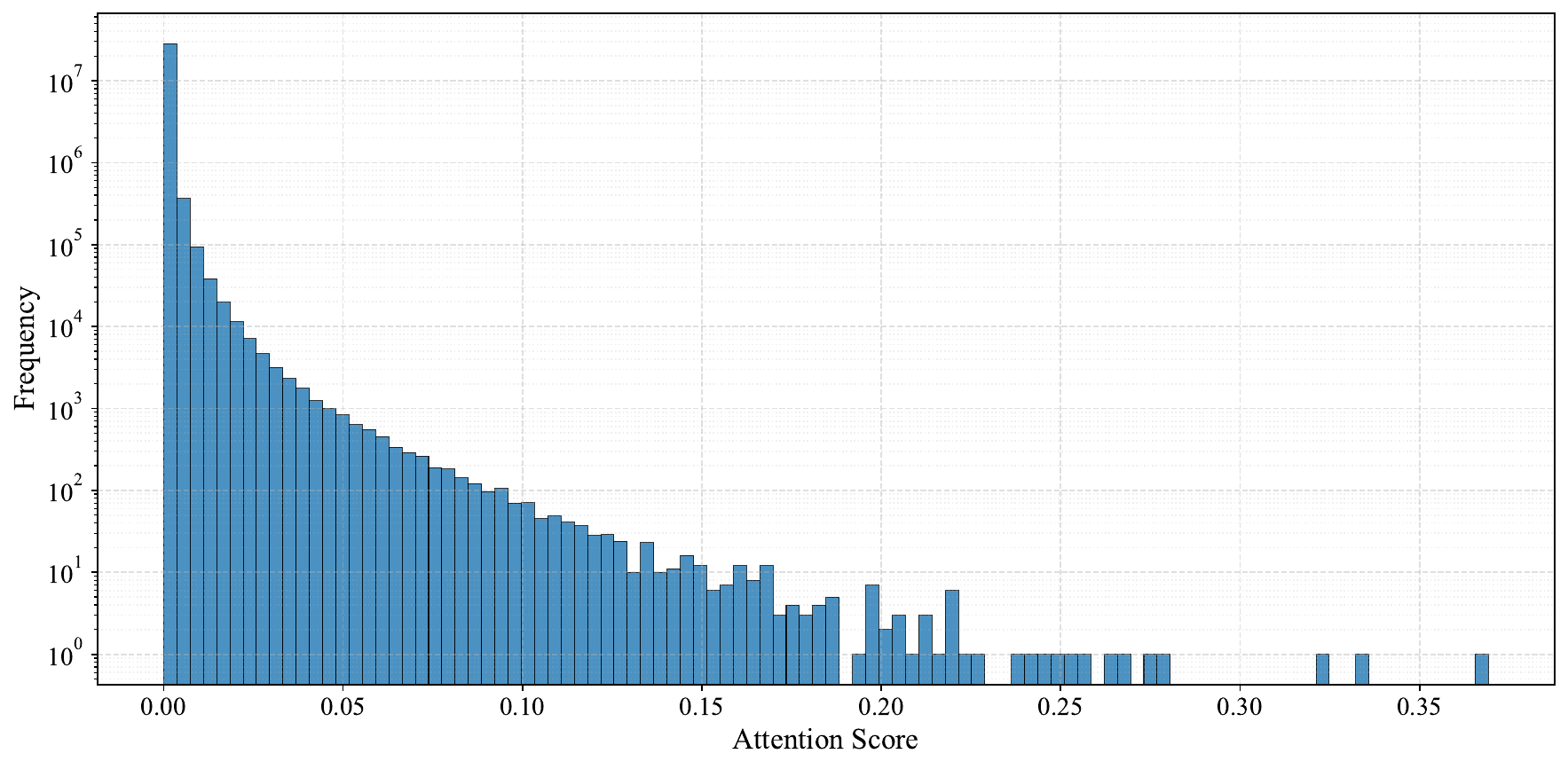}
   \caption{Long-tailed distribution of attention scores. Valid correspondences appear as sparse, high-value outliers against dominant background noise.
  }
   \label{fig:attn}
\end{figure}

Standard cross-attention mechanisms are prone to semantic leakage in asymmetric settings. While recent methods like CACTIF\cite{CACTIF} filter correspondences using a fixed percentile of feature similarity, this imposes a rigid assumption of a constant signal-to-noise ratio across all images, layers, and timesteps. To dynamically isolate valid style features, an adaptive thresholding strategy is essential.

A straightforward strategy is to define this boundary using standard parametric statistics, such as the mean and standard deviation. However, visualizing the cross-attention affinity matrix (\cref{fig:attn}) reveals a severe heavy-tailed distribution: valid structural correspondences manifest as extremely sparse, high-value outliers against dominant background noise. In such highly skewed data, these outliers severely bias the mean and inflate the variance, rendering standard statistics unreliable. To rigorously address this, we propose a dual-constrained filtering mechanism, prioritizing a robust statistical metric followed by an explicit semantic barrier.

Let $\mathbf{A}$ denote the raw cross-attention map. Instead of the conventional mean and standard deviation, we utilize the robust median and Median Absolute Deviation (MAD). This robust thresholding strategy is designed to suppress the influence of rare extreme attention responses while retaining reliable style correspondences. Specifically, the MAD is defined as

\begin{equation}
\text{MAD}(\mathbf{A}) = \text{Median}(|\mathbf{A} - \text{Median}(\mathbf{A})|)
\end{equation}

Then, the decision boundary $\tau$ and the resulting statistical mask $\mathcal{M}_{stat}$ are computed as follows
\begin{equation}
\tau = \text{Median}(\mathbf{A}) + \lambda \cdot \text{MAD}(\mathbf{A})
\end{equation}

\begin{equation}
  \mathcal{M}_{stat}(i, j) = \mathbb{I}[\mathbf{A}_{i,j} > \tau]
\end{equation}

While prior works like CACTIF\cite{CACTIF} rely entirely on feature similarities and lack explicit semantic boundaries during filtering, this inevitably leads to cross-class texture pollution. To prohibit this, we construct a binary semantic mask $\mathcal{M}_{sem}$ leveraging the refined pseudo-label $\hat{M}_{s}^t$ from \cref{eq:mask}

\begin{equation}
  \mathcal{M}_{sem}(i, j) = \mathbb{I}[M_c(i) = \hat{M}_{s}(j)] \cdot \mathbb{I}[\hat{M}_{s}(j) \neq \emptyset]
\end{equation}

The final attention filter is the intersection
\begin{equation}
\mathcal{M} = \mathcal{M}_{sem} \odot \mathcal{M}_{stat}
\end{equation}

To prevent feature collapse and preserve structural integrity, we implement a soft-gated fallback mechanism. We compute the filtered style features $\mathbf{h}^{style}_i$, where $\mathbf{h}$ denotes the hidden state of the attention layer, and a spatial preservation score $\alpha_i \in [0, 1]$, which quantifies the reliability of the injected style

\begin{equation}
\mathbf{h}^{style}_i = \sum_{j} \left( \mathcal{M}(i,j) \cdot \mathbf{A}(i,j) \right) \mathbf{v}^{style}_j
\end{equation}

\begin{equation}
\alpha_i = \sum_{j} \mathcal{M}(i,j) \cdot \mathbf{A}(i,j)
\label{eq:alpha}
\end{equation}

The final hidden state $\mathbf{h}_i$ is then synthesized via adaptive mixing

\begin{equation}
\mathbf{h}_i = \mathbf{h}^{style}_i + (1 - \alpha_i) \cdot \mathbf{h}^{content}_i
\end{equation}
where $\mathbf{h}^{content}_i$ denotes the original output of the content-branch attention.

This dynamic formulation ensures the model prioritizes stylized textures $\mathbf{h}^{style}$ in high-confidence regions ($\alpha_i \to 1$), while reverting to the original content structure $\mathbf{h}^{content}$ where correspondences are ambiguous or absent ($\alpha_i \to 0$).

\subsection{Pixel-Proportion Modulated Hybrid AdaIN}
\label{subsec:adain}
Standard Class-wise AdaIN\cite{CACTIF} relies on the ideal assumption that pixel-perfect semantic annotations are available for both domains. However, this assumption collapses in our asymmetric setting, where the style segmentation $\hat{M}_{s}$ is inferred and inevitably noisy. Consequently, a specific semantic class might occupy a significant portion of the synthetic content image but appear only as a few fragmented pixels or be entirely absent in the single real-world style reference. Relying on such noisy statistics to normalize a large content region can lead to catastrophic feature distortion, where the style of a few outliers dominates the appearance of the entire object.

To mitigate this, we propose a robust normalization scheme that dynamically calibrates the strength of style injection. By explicitly tying the reliability of feature statistics to their relative pixel abundance, it prevents blind statistical alignment onto unreliable regions.

For each semantic class $k \in \{0, \dots, C-1\}$, we first compute the channel-wise mean and standard deviation for both the content features $\mathbf{F}_{c}$ (denoted as $\mu_{content}^{(k)}$, $\sigma_{content}^{(k)}$) and the style features $\mathbf{F}_{s}$ (denoted as $\mu_{style}^{(k)}$, $\sigma_{style}^{(k)}$) within their respective masks, $M_{c}^{(k)}$ and $\hat{M}_{s}^{(k)}$. Let $N_{c}^k = |M_{c}^{(k)}|$ and $N_{s}^k = |\hat{M}_{s}^{(k)}|$ denote the number of pixels for class $k$ in the content and style masks, respectively. We define the relative difference ratio 
\begin{equation}
  \delta_k = \frac{N_{c}^k - N_{s}^k}{N_{c}^k + N_{s}^k}
\end{equation}

The modulation factor is computed via a sigmoid-based confidence function
\begin{equation}
  \beta_k = Sigmoid\left( \gamma \cdot \delta_k \right)
\end{equation}
where $\gamma$ is a scaling hyperparameter.

The final stylized feature $\mathbf{F}_{out}$ is synthesized by linearly interpolating between original content features and AdaIN-transformed features, weighted by the class-specific modulation map $\mathbf{B}$, where $\mathbf{B}(i,j) = \beta_k$ for pixels belonging to class $k$.
\begin{equation}
  \mathbf{F}_{aligned} = \sigma_{style}^{k} \left( \frac{\mathbf{F}_{c} - \mu_{content}^{k}}{\sigma_{content}^{k}} \right) + \mu_{style}^{k}
\end{equation}

\begin{equation}
  \mathbf{F}_{out} = \mathbf{B} \odot \mathbf{F}_{c} + (\mathbf{1} - \mathbf{B}) \odot \mathbf{F}_{aligned}
\end{equation}

This pixel-proportion modulation  ensures that style transfer is aggressive only where the semantic evidence is abundant and trustworthy, while gracefully degrading to content preservation in ambiguous or sparse regions.

\section{Experiments}
\label{sec:exp}
\subsection{Experimental Setup}
Following established protocols in prior literature\cite{CACTIF}, we utilize 5,000 annotated images from the GTA dataset\cite{gta} as the synthetic source. For the target domain, we evaluate our approach under an asymmetric setting using a single unlabeled reference image. As a representative case study, the main text focuses on the "Vegetation Dominant" scene (\cref{fig:tu1}, upper-right). The robust thresholding parameter $\lambda$ is set to 2.0, and the fusion scale $\gamma$ is set to 8.0. Evaluation relies on SegFormer\cite{segformer} for segmentation and LPIPS\cite{lpips} for structural fidelity. All experiments are conducted on a single NVIDIA RTX 4090 GPU. We benchmark ASTModel against training-free diffusion-based method: Cross-Image Attention\cite{crossimage} and CACTIF\cite{CACTIF}. Comprehensive experimental details and results across diverse target scenarios are provided in the Supplementary Material.

\subsection{Qualitative Evaluation}

\begin{figure}[t]
  \centering
 \includegraphics[width=0.95\linewidth]{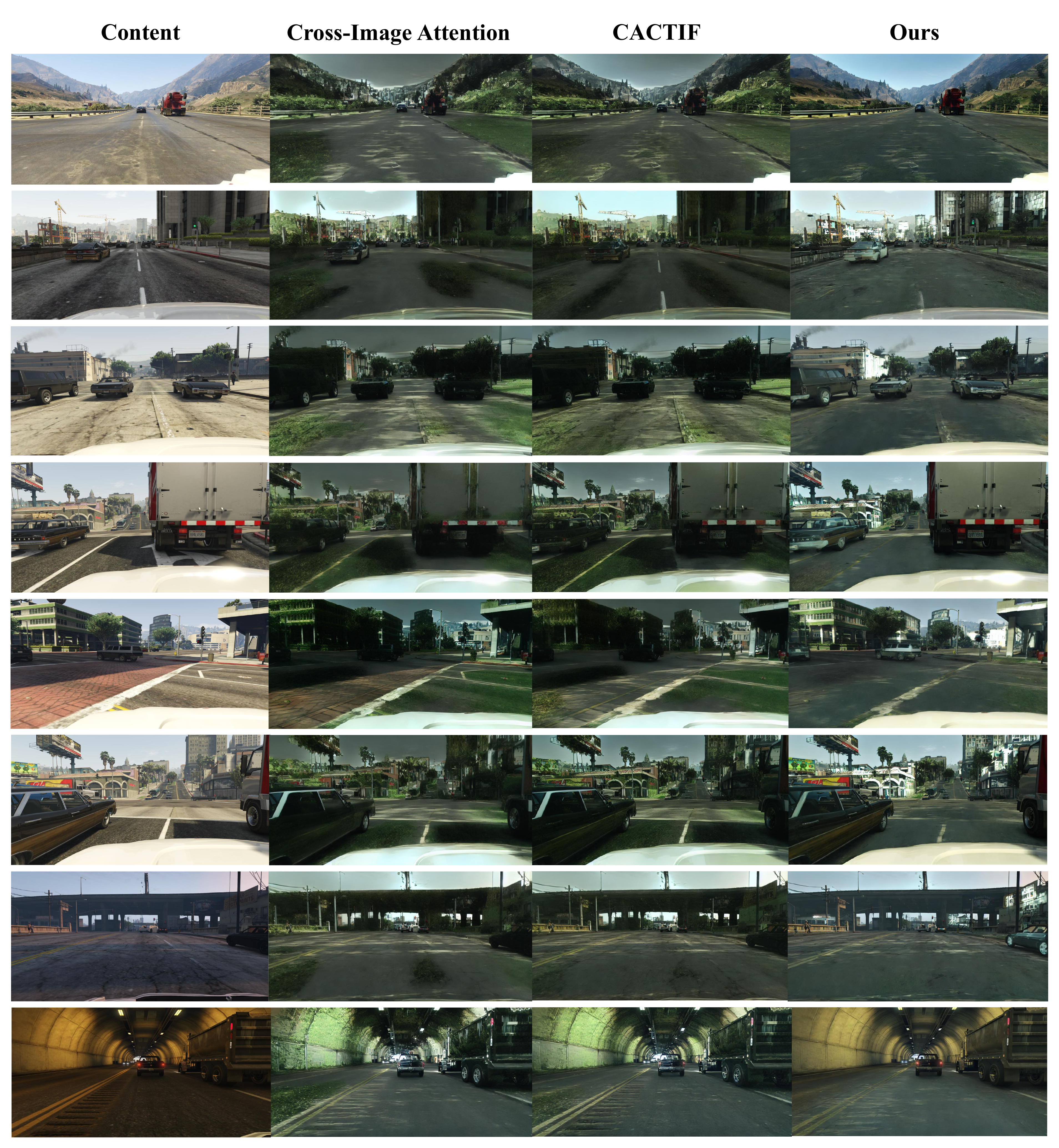}
   \caption{Qualitative comparison of ASTModel and baseline methods.
  }
   \label{fig:result_grass}
\end{figure}

\subsubsection{Accurate Semantic Alignment}
As shown in the first row of \cref{fig:result_grass}, baselines erroneously map "mountain" regions to "building" textures, as they rely on superficial appearance similarity while overlooking semantic distinction. Conversely, ASTModel preserves semantic identity through Prototype-Guided Style Semantic Prior Extraction and Multi-Layer Semantic Voting. By anchoring semantic prototypes with foundation model priors and exploiting cross-layer diffusion consistency, our framework ensures content queries attend purely to semantically congruent style keys.

\subsubsection{Mitigating Semantic Leakage}
 In \cref{fig:result_grass}, baselines suffer severe semantic leakage by projecting dominant greenery onto roads and vehicles. ASTModel mitigates this via two mechanisms: Semantically Constrained Adaptive Attention Filtering enforces strict semantic barriers for texture injection, while Pixel-Proportion Modulated Hybrid AdaIN calibrates stylization intensity based on the statistical reliability of each class. It prevents overfitting to dominant textures, preserving the semantic purity in asymmetric settings.

\subsubsection{High-Fidelity Structural Preservation}
As shown in the third row of \cref{fig:result_grass}, baselines suffer from significant detail loss, often reducing vehicles to blurred shapes. Conversely, ASTModel preserves high-fidelity structures, maintaining crisp contours and fine-grained textures for intricate objects like vehicle outlines and architectural elements. This superior clarity ensures that the generated datasets serve as high-quality training data for downstream perception tasks.

\subsubsection{Robustness to Reference-Unseen Scenarios}
In challenging scenarios like "Tunnel" and "Sunset" (bottom two rows of \cref{fig:result_grass}), baselines struggle with absent style references, generating noise by forcing invalid correspondences. ASTModel maintains structural fidelity by detecting these weak signals: Attention Filtering reverts to content features when correspondences drop, and Pixel-Proportion Modulated Hybrid AdaIN suppresses injection for statistically sparse features. This dynamic adaptability ensures that ASTModel can gracefully handle a wide range of target styles, even those with minimal semantic overlap with the source domain.
\subsection{Quantitative Evaluation}

\subsubsection{Downstream Perception Utility}

\begin{table}[t]
\centering
\begin{minipage}[t]{0.58\textwidth}
\centering
\caption{Quantitative Evaluation on Downstream Segmentation}
\label{tab:quant_seg}
\begin{tabular}{lcc}
\toprule
\textbf{Method} & \textbf{Pixel Acc} $\uparrow$ & \textbf{mIoU} $\uparrow$ \\
\midrule
Source Only\cite{gta} & 0.844 & 0.275\\
\midrule
Cross-Image Attn.\cite{crossimage} & 0.653 & 0.224 \\
CACTIF\cite{CACTIF} & 0.782 & 0.289 \\
\textbf{ASTModel (Ours)} & \textbf{0.847} & \textbf{0.309} \\
\bottomrule
\end{tabular}
\end{minipage}\hfill
\begin{minipage}[t]{0.4\textwidth}
\centering
\caption{Quantitative Evaluation on Structural Fidelity}
\label{tab:quant_struct}
\begin{tabular}{lc}
\toprule
\textbf{Method} & \textbf{LPIPS} $\downarrow$ \\
\midrule
Cross-Image Attn.\cite{crossimage} & 0.5205 \\
CACTIF\cite{CACTIF} & 0.4184 \\
\textbf{ASTModel (Ours)} & \textbf{0.3588} \\
\bottomrule
\end{tabular}
\end{minipage}
\end{table}

\begin{table}[t]
\centering
\begin{minipage}[t]{0.52\textwidth}
\centering
\caption{Computational Efficiency Comparison. Total refers to the entire 5,000 images.}
\label{tab:quant_time}
\begin{tabular}{l|cc}
\toprule
\textbf{Method} & \textbf{Time} $\downarrow$ & \textbf{Total} $\downarrow$ \\
\midrule
Cross-Image Attn.\cite{crossimage} & 18s & 28h \\
CACTIF\cite{CACTIF} & 80s & 114h \\
ASTModel (Ours) & 25s & 37h \\
\bottomrule
\end{tabular}
\end{minipage}\hfill
\begin{minipage}[t]{0.48\textwidth}
\centering
\caption{Quantitative Ablation Study}
\label{tab:ablation_study}
\begin{tabular}{l|c}
\toprule
\textbf{Method} & \textbf{LPIPS ↓} \\ 
\midrule
w/o Attention Filtering & 0.5114 \\
w/o Hybrid AdaIN        & 0.3737 \\
w/o Semantic Voting     & 0.3641 \\
\midrule
\textbf{Ours (Full Model)} & \textbf{0.3588} \\
\bottomrule
\end{tabular}
\label{tab:ablation}
\end{minipage}
\end{table}

The primary metric for synthetic-to-real adaptation is its impact on downstream perception tasks. As shown in \cref{tab:quant_seg}, our ASTModel consistently achieves the highest Pixel Accuracy and mIoU, demonstrating its superior ability to effectively transfer target styles while strictly preserving original semantic contents compared to all baselines. Further details on the segmentation model's training and validation are provided in the Supplementary Material.
\subsubsection{Structural Fidelity}
Beyond semantic correctness, autonomous driving applications mandate the strict preservation of geometric layouts and structural boundaries. \cref{tab:quant_struct} demonstrates that ASTModel achieves the lowest LPIPS score, indicating minimal structural distortion. The baselines exhibit higher LPIPS scores due to severe spatial warping and hallucination artifacts in the asymmetric setting. 
\subsubsection{Computational Efficiency}
\cref{tab:quant_time} details the inference latency for synthesizing the 5,000-image dataset. CACTIF suffers from a severe computational bottleneck (80s/image, 114h total) due to exhaustive pixel-wise feature similarity calculations in its attention filtering. In contrast, ASTModel achieves a 3.2$\times$ speedup (25s/image). This high efficiency is achieved by replacing computationally heavy dense matching with robust statistical metrics to establish adaptive decision boundaries.

\subsection{Ablation Study}
\begin{figure}[t]
  \centering
 \includegraphics[width=1.0\linewidth]{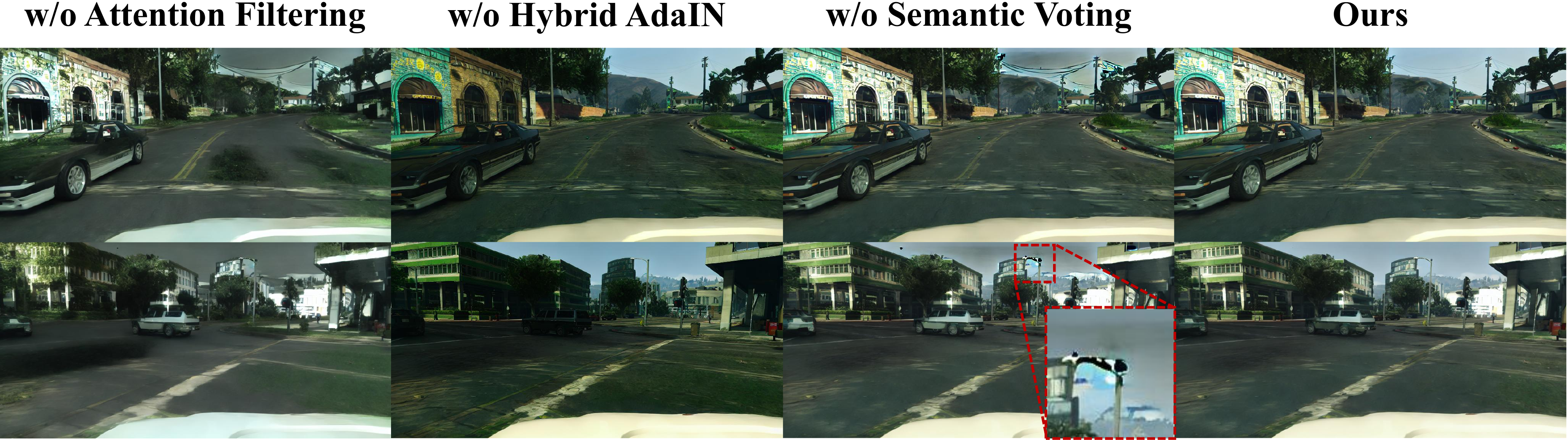}
   \caption{Ablation study on key components of ASTModel.
  }
   \label{fig:abl}
\end{figure}

\subsubsection{Effectiveness of Semantically Constrained Adaptive Attention Filtering}
Without filtering, the model incorrectly maps texture from the vegetation-heavy style reference onto shadowed road regions in the content image (\cref{fig:abl}). This results in cross-class pollution, manifested as unnatural greenish, moss-like textures on the road, spiking LPIPS to 0.5114 (\cref{tab:ablation}). Our semantic barrier successfully restricts style injection exclusively to congruent categories.
\subsubsection{Effectiveness of Pixel-Proportion Modulated Hybrid AdaIN}
Reverting to global AdaIN blindly forces dominant style statistics onto the entire image, causing severe color casts and identity loss (\eg, the darkened SUV in \cref{fig:abl}, Row 2). Additionally, the overall scene suffers from an unnatural global color cast. Our hybrid approach aligns statistics within corresponding categories and scales injection strength based on pixel abundance.
\subsubsection{Effectiveness of Multi-Layer Semantic Voting}
Without this voting mechanism, the coarse DINO prior fails to precisely isolate small, irregular regions in the style reference, such as fragmented skies or thin utility poles. Consequently, mismatched style features are transferred: the sky region is corrupted by unrelated textures and artifacts (\cref{fig:abl}, Row 1), and poles suffer from feature loss, resulting in disconnected or blurred structures (\cref{fig:abl}, Row 2). The quantitative benefit of correcting these artifacts is evident in \cref{tab:ablation}. By aggregating consensus from the layers of the diffusion decoder, our mechanism refines these coarse priors and preserves intricate details.

\subsection{Pseudo-Label Refinement Validation}
\begin{figure}[t]
\centering
\begin{minipage}{0.65\linewidth}
\centering
\includegraphics[width=\linewidth]{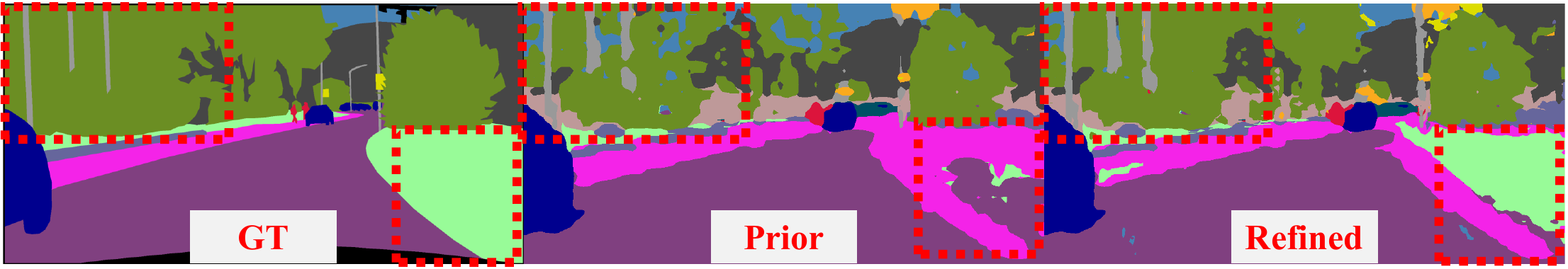}
\caption{Validation of pseudo-label refinement.}
\label{fig:pseudo_validation}
\end{minipage}\hfill
\begin{minipage}{0.3\linewidth}
\centering
\includegraphics[width=0.8\linewidth]{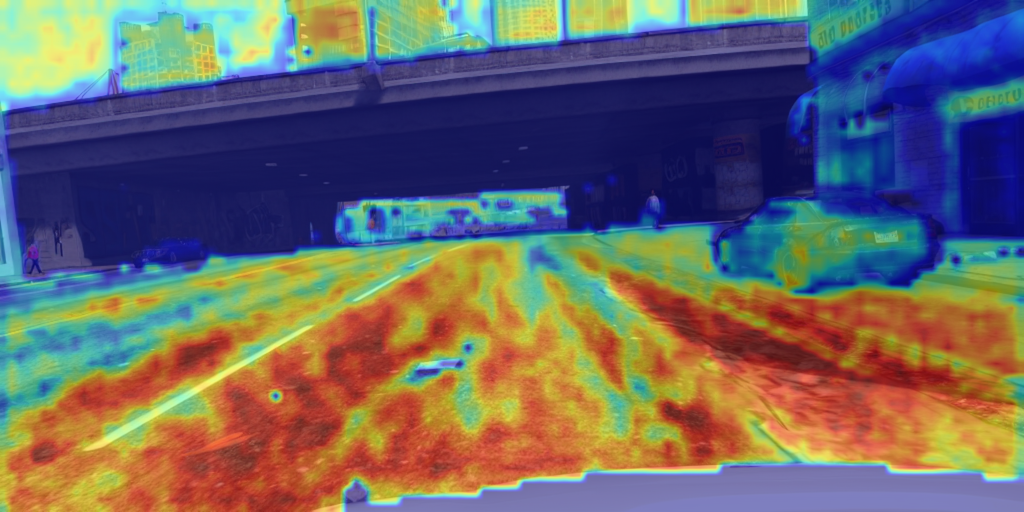}
\caption{Visualization of $\alpha$.}
\label{fig:alpha_visualization}
\end{minipage}
\end{figure}

We further evaluate the quality of the style-side pseudo labels. As shown in \cref{fig:pseudo_validation}, the Stage-I prior captures coarse semantics but contains inaccurate boundaries and local mismatches. Stage-II refinement improves boundary alignment and corrects ambiguous regions. Quantitatively, Pixel Accuracy increases from $0.688$ to $0.790$, and mIoU improves from $0.281$ to $0.339$, verifying that multi-layer semantic voting provides more reliable guidance for class-consistent style injection. Notably, the refinement is adaptive to content-style semantic compatibility: stronger semantic alignment between the content image and style reference leads to more consistent multi-layer voting, more accurate pseudo labels, and consequently higher-quality style transfer.

\subsection{Visualization of Preservation Score} 
\label{sec:alpha_visualization}

We visualize the preservation score $\alpha$ in \cref{eq:alpha} to analyze how ASTModel balances style injection and content preservation. As shown in \cref{fig:alpha_visualization}, regions with reliable semantic correspondences obtain higher $\alpha$ and receive stronger style transfer. In contrast, regions that are absent or weakly represented in the style image obtain lower $\alpha$, causing the model to fall back to content preservation. For example, the bridge region has low $\alpha$ because no corresponding bridge appears in the style image, which prevents invalid texture injection and reduces semantic leakage.

\subsection{Robustness of Median/MAD Filtering}
\label{sec:mad_analysis}

\begin{figure}[t]
\centering
\begin{minipage}{0.4\linewidth}
\centering
\captionof{table}{Robustness comparison. Values are scaled by $10^{-5}$.}
\label{tab:mad_analysis}
\begin{tabular}{@{}lcc@{}}
\hline
\textbf{Stat.}  & \textbf{Orig.} & \textbf{Top-rm.} \\
\hline
Mean/Std & 41.8/173.3 & 41.8/171.9 \\
Med./MAD & 5.1/4.8 & 5.1/4.8 \\
\hline
\end{tabular}
\end{minipage}\hfill
\begin{minipage}{0.55\linewidth}
\centering
\includegraphics[width=\linewidth]{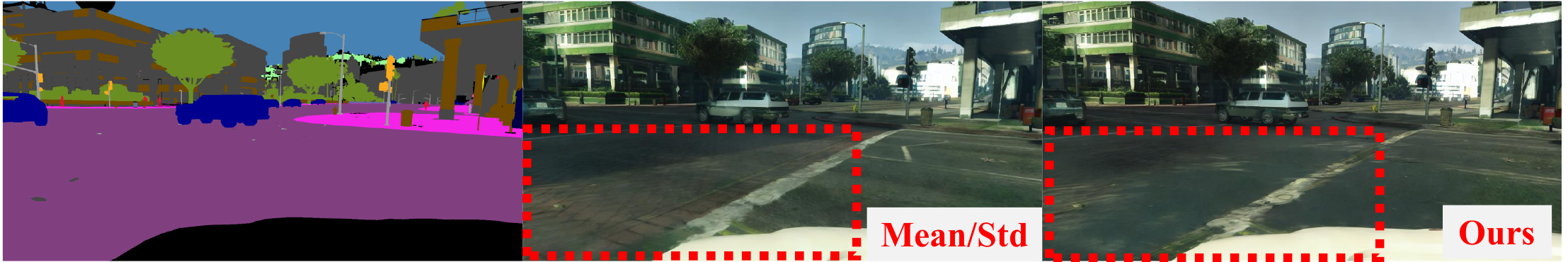}
\caption{Visual comparison between Mean/Std and Median/MAD filtering.}
\label{fig:mad_analysis}
\end{minipage}
\end{figure}

We further validate the proposed Median/MAD threshold against a Mean/Std-based alternative. As shown in \cref{tab:mad_analysis}, removing the top-20 largest attention scores changes the Std-based scale from $173.3$ to $171.9$, while the Median/MAD statistics remain unchanged. Mean/Std filtering preserves only $1.8\%$ of attention entries, leading to over-filtering and insufficient style transfer. In contrast, Median/MAD preserves $33.2\%$ of entries and retains more reliable correspondences under the heavy-tailed attention distribution. This difference is also reflected in \cref{fig:mad_analysis}: Mean/Std under-transfers the road region and leaves source-domain textures insufficiently adapted, whereas Median/MAD produces more complete and natural road-style transfer.

\subsection{Discussion}
\subsubsection{Scalability with Foundation Models}
As a modular, training-free framework, ASTModel is inherently scalable. Future upgrades to vision encoder or diffusion backbone will translate to enhanced semantic alignment and photorealism.
\subsubsection{Robustness Across Domains}
Conventional adaptation methods require per-domain retraining. ASTModel derives its robust generalizability fundamentally from its reference-guided style transfer paradigm. This enables our framework to seamlessly tackle other critical distribution shifts, such as severe weather and day-to-night translation. Results are provided in the Supplementary Material.

\section{Conclusion}
In this paper, we introduced Asymmetric Style Transfer for Autonomous Driving (ASTAD), a task formulation that addresses the practical setting where dense semantic labels are available for synthetic source data but absent for real-world style references. To tackle this challenge, we proposed ASTModel, a training-free framework designed to bridge the synthetic-to-real gap under asymmetric conditions. Experiments demonstrate that ASTModel improves semantic consistency and inference efficiency over relevant training-free baselines under the evaluated protocol.

A limitation of this work is that the current protocol relies on a single reference image, whose scene coverage is inherently limited. As shown in \cref{fig:alpha_visualization}, for example, the rear part of the car receives a stronger response than the front because car-related cues in the reference are spatially imbalanced and dominated by rear-side regions. Future work will extend ASTAD to multiple reference images, which can provide more balanced category-level style statistics, cover sparse local cues more comprehensively, and reduce the conflict between semantic safety and style richness in challenging scenes. By transforming label-rich synthetic data into semantically reliable target-style samples, ASTAD offers a practical path toward scalable and robust perception model development for autonomous driving.

%
%
\bibliographystyle{splncs04}
\bibliography{main}

@String(CVPR  = {IEEE Conf. Comput. Vis. Pattern Recog.})

@String(ECCV  = {Eur. Conf. Comput. Vis.})

@String(CVPR  = {CVPR})

@String(ECCV  = {ECCV})

@inproceedings{cityscapes,
  title={The cityscapes dataset for semantic urban scene understanding},
  author={Cordts, Marius and Omran, Mohamed and Ramos, Sebastian and Rehfeld, Timo and Enzweiler, Markus and Benenson, Rodrigo and Franke, Uwe and Roth, Stefan and Schiele, Bernt},
  booktitle={Proceedings of the IEEE conference on computer vision and pattern recognition},
  pages={3213--3223},
  year={2016}
}

@inproceedings{2,
  title={Daformer: Improving network architectures and training strategies for domain-adaptive semantic segmentation},
  author={Hoyer, Lukas and Dai, Dengxin and Van Gool, Luc},
  booktitle={Proceedings of the IEEE/CVF conference on computer vision and pattern recognition},
  pages={9924--9935},
  year={2022}
}

@article{3,
  title={Synthetic datasets for autonomous driving: A survey},
  author={Song, Zhihang and He, Zimin and Li, Xingyu and Ma, Qiming and Ming, Ruibo and Mao, Zhiqi and Pei, Huaxin and Peng, Lihui and Hu, Jianming and Yao, Danya and others},
  journal={IEEE Transactions on Intelligent Vehicles},
  volume={9},
  number={1},
  pages={1847--1864},
  year={2023},
  publisher={IEEE}
}

@inproceedings{gta,
  title={Playing for data: Ground truth from computer games},
  author={Richter, Stephan R and Vineet, Vibhav and Roth, Stefan and Koltun, Vladlen},
  booktitle={European conference on computer vision},
  pages={102--118},
  year={2016},
  organization={Springer}
}

@article{5,
  title={A Style-Based Profiling Framework for Quantifying the Synthetic-to-Real Gap in Autonomous Driving Datasets},
  author={Yao, Dingyi and Han, Xinyao and Ming, Ruibo and Song, Zhihang and Peng, Lihui and Hu, Jianming and Yao, Danya and Zhang, Yi},
  journal={arXiv preprint arXiv:2510.10203},
  year={2025}
}

@article{6,
  title={Synthetic dataset evaluation based on generalized cross validation},
  author={Song, Zhihang and Yao, Dingyi and Ming, Ruibo and Peng, Lihui and Yao, Danya and Zhang, Yi},
  journal={arXiv preprint arXiv:2509.11273},
  year={2025}
}

@article{diffusion,
  title={Denoising diffusion implicit models},
  author={Song, Jiaming and Meng, Chenlin and Ermon, Stefano},
  journal={arXiv preprint arXiv:2010.02502},
  year={2020}
}

@inproceedings{ldm,
  author       = {Robin Rombach and
                  Andreas Blattmann and
                  Dominik Lorenz and
                  Patrick Esser and
                  Bj{\"{o}}rn Ommer},
  title        = {High-Resolution Image Synthesis with Latent Diffusion Models},
  booktitle    = {{IEEE/CVF} Conference on Computer Vision and Pattern Recognition,
                  {CVPR} 2022, New Orleans, LA, USA, June 18-24, 2022},
  pages        = {10674--10685},
  publisher    = {{IEEE}},
  year         = {2022}
}

@article{tcvst4,
  title={Weafu: Weather-informed image blind restoration via multi-weather distribution diffusion},
  author={Cheng, Bodong and Li, Juncheng and Shi, Jun and Fang, Yingying and Zhang, Guixu and Chen, Yin and Zeng, Tieyong and Li, Zhi},
  journal={IEEE Transactions on Circuits and Systems for Video Technology},
  year={2024},
  publisher={IEEE}
}

@article{tcvst5,
  title={Toward high-quality HDR deghosting with conditional diffusion models},
  author={Yan, Qingsen and Hu, Tao and Sun, Yuan and Tang, Hao and Zhu, Yu and Dong, Wei and Van Gool, Luc and Zhang, Yanning},
  journal={IEEE Transactions on Circuits and Systems for Video Technology},
  volume={34},
  number={5},
  pages={4011--4026},
  year={2023},
  publisher={IEEE}
}

@article{tcvst6,
  title={Efficient image enhancement with a diffusion-based frequency prior},
  author={Yan, Qingsen and Hu, Tao and Wu, Peng and Dai, Duwei and Gu, Shuhang and Dong, Wei and Zhang, Yanning},
  journal={IEEE Transactions on Circuits and Systems for Video Technology},
  year={2025},
  publisher={IEEE}
}

@inproceedings{crossimage,
  title={Cross-image attention for zero-shot appearance transfer},
  author={Alaluf, Yuval and Garibi, Daniel and Patashnik, Or and Averbuch-Elor, Hadar and Cohen-Or, Daniel},
  booktitle={ACM SIGGRAPH 2024 conference papers},
  pages={1--12},
  year={2024}
}

@article{zstar,
  title={ Z$^{*} $: Zero-shot Style Transfer via Attention Rearrangement},
  author={Deng, Yingying and He, Xiangyu and Tang, Fan and Dong, Weiming},
  journal={arXiv preprint arXiv:2311.16491},
  year={2023}
}

@inproceedings{styleid,
  title={Style injection in diffusion: A training-free approach for adapting large-scale diffusion models for style transfer},
  author={Chung, Jiwoo and Hyun, Sangeek and Heo, Jae-Pil},
  booktitle={Proceedings of the IEEE/CVF conference on computer vision and pattern recognition},
  pages={8795--8805},
  year={2024}
}

@article{eyeforaneye,
  title={Eye-for-an-eye: Appearance Transfer with Dense Semantic Correspondence in Diffusion Models},
  author={Go, Sooyeon and Choi, Kyungmook and Shin, Minjung and Uh, Youngjung},
  journal={Proceedings of the IEEE/CVF Winter Conference on Applications of Computer Vision},
  pages={4641--4650},
  year={2026}
}

@article{CACTIF,
  title={Style transfer with diffusion models for synthetic-to-real domain adaptation},
  author={Chigot, Estelle and Wilson, Dennis G and Ghrib, Meriem and Oberlin, Thomas},
  journal={Computer Vision and Image Understanding},
  pages={104445},
  year={2025},
  publisher={Elsevier}
}

@inproceedings{rw1,
  title={Image style transfer using convolutional neural networks},
  author={Gatys, Leon A and Ecker, Alexander S and Bethge, Matthias},
  booktitle={Proceedings of the IEEE conference on computer vision and pattern recognition},
  pages={2414--2423},
  year={2016}
}

@inproceedings{rw2,
  title={Arbitrary style transfer in real-time with adaptive instance normalization},
  author={Huang, Xun and Belongie, Serge},
  booktitle={Proceedings of the IEEE international conference on computer vision},
  pages={1501--1510},
  year={2017}
}

@article{tcvst8,
  title={Adaptive conditional denoising diffusion model with hybrid affinity regularizer for generalized zero-shot learning},
  author={Gao, Mengyu and Dong, Qiulei},
  journal={IEEE Transactions on Circuits and Systems for Video Technology},
  volume={34},
  number={7},
  pages={5641--5652},
  year={2024},
  publisher={IEEE}
}

@inproceedings{rw4,
  title={Image-to-image translation with conditional adversarial networks},
  author={Isola, Phillip and Zhu, Jun-Yan and Zhou, Tinghui and Efros, Alexei A},
  booktitle={Proceedings of the IEEE conference on computer vision and pattern recognition},
  pages={1125--1134},
  year={2017}
}

@inproceedings{rw5,
  title={Unpaired image-to-image translation using cycle-consistent adversarial networks},
  author={Zhu, Jun-Yan and Park, Taesung and Isola, Phillip and Efros, Alexei A},
  booktitle={Proceedings of the IEEE international conference on computer vision},
  pages={2223--2232},
  year={2017}
}

@inproceedings{rw5-2,
  title={Contrastive learning for unpaired image-to-image translation},
  author={Park, Taesung and Efros, Alexei A and Zhang, Richard and Zhu, Jun-Yan},
  booktitle={European conference on computer vision},
  pages={319--345},
  year={2020},
  organization={Springer}
}

@inproceedings{rw6,
  title={Multimodal unsupervised image-to-image translation},
  author={Huang, Xun and Liu, Ming-Yu and Belongie, Serge and Kautz, Jan},
  booktitle={Proceedings of the European conference on computer vision (ECCV)},
  pages={172--189},
  year={2018}
}

@inproceedings{rw7-1,
  title={Stargan: Unified generative adversarial networks for multi-domain image-to-image translation},
  author={Choi, Yunjey and Choi, Minje and Kim, Munyoung and Ha, Jung-Woo and Kim, Sunghun and Choo, Jaegul},
  booktitle={Proceedings of the IEEE conference on computer vision and pattern recognition},
  pages={8789--8797},
  year={2018}
}

@inproceedings{rw7-2,
  title={Stargan v2: Diverse image synthesis for multiple domains},
  author={Choi, Yunjey and Uh, Youngjung and Yoo, Jaejun and Ha, Jung-Woo},
  booktitle={Proceedings of the IEEE/CVF conference on computer vision and pattern recognition},
  pages={8188--8197},
  year={2020}
}

@inproceedings{stylediffusion,
  title={Stylediffusion: Controllable disentangled style transfer via diffusion models},
  author={Wang, Zhizhong and Zhao, Lei and Xing, Wei},
  booktitle={Proceedings of the IEEE/CVF international conference on computer vision},
  pages={7677--7689},
  year={2023}
}

@inproceedings{blora,
  title={Implicit style-content separation using b-lora},
  author={Frenkel, Yarden and Vinker, Yael and Shamir, Ariel and Cohen-Or, Daniel},
  booktitle={European Conference on Computer Vision},
  pages={181--198},
  year={2024},
  organization={Springer}
}

@inproceedings{cycada,
  title={Cycada: Cycle-consistent adversarial domain adaptation},
  author={Hoffman, Judy and Tzeng, Eric and Park, Taesung and Zhu, Jun-Yan and Isola, Phillip and Saenko, Kate and Efros, Alexei and Darrell, Trevor},
  booktitle={International conference on machine learning},
  pages={1989--1998},
  year={2018},
  organization={Pmlr}
}

@inproceedings{shade,
  title={Style-hallucinated dual consistency learning for domain generalized semantic segmentation},
  author={Zhao, Yuyang and Zhong, Zhun and Zhao, Na and Sebe, Nicu and Lee, Gim Hee},
  booktitle={European conference on computer vision},
  pages={535--552},
  year={2022},
  organization={Springer}
}

@inproceedings{dginstyle,
  title={Dginstyle: Domain-generalizable semantic segmentation with image diffusion models and stylized semantic control},
  author={Jia, Yuru and Hoyer, Lukas and Huang, Shengyu and Wang, Tianfu and Van Gool, Luc and Schindler, Konrad and Obukhov, Anton},
  booktitle={European Conference on Computer Vision},
  pages={91--109},
  year={2024},
  organization={Springer}
}

@article{simgen,
  title={Simgen: Simulator-conditioned driving scene generation},
  author={Zhou, Yunsong and Simon, Michael and Peng, Zhenghao and Mo, Sicheng and Zhu, Hongzi and Guo, Minyi and Zhou, Bolei},
  journal={Advances in Neural Information Processing Systems},
  volume={37},
  pages={48838--48874},
  year={2024}
}

@inproceedings{weatherdiff,
  title={Weather-Diff: Towards Arbitrary Adversarial Weather Generation with Diffusion Models},
  author={Wang, Zeda and Gao, Huan-ang and Zhang, Guiyu and Zhao, Hao},
  booktitle={International Conference on Intelligent Computing},
  pages={134--145},
  year={2025},
  organization={Springer}
}

@article{dinov2,
  title={DINOv2: Learning Robust Visual Features without Supervision},
  author={Oquab, Maxime and Darcet, Timoth{\'e}e and Moutakanni, Th{\'e}o and Vo, Huy and Szafraniec, Marc and Khalidov, Vasil and Fernandez, Pierre and Haziza, Daniel and Massa, Francisco and El-Nouby, Alaaeldin and others},
  journal={Transactions on Machine Learning Research Journal},
  year={2024}
}

@article{segformer,
  title={SegFormer: Simple and efficient design for semantic segmentation with transformers},
  author={Xie, Enze and Wang, Wenhai and Yu, Zhiding and Anandkumar, Anima and Alvarez, Jose M and Luo, Ping},
  journal={Advances in neural information processing systems},
  volume={34},
  pages={12077--12090},
  year={2021}
}

@inproceedings{lpips,
  title={The unreasonable effectiveness of deep features as a perceptual metric},
  author={Zhang, Richard and Isola, Phillip and Efros, Alexei A and Shechtman, Eli and Wang, Oliver},
  booktitle={Proceedings of the IEEE conference on computer vision and pattern recognition},
  pages={586--595},
  year={2018}
}

@inproceedings{adaattn,
  title={Adaattn: Revisit attention mechanism in arbitrary neural style transfer},
  author={Liu, Songhua and Lin, Tianwei and He, Dongliang and Li, Fu and Wang, Meiling and Li, Xin and Sun, Zhengxing and Li, Qian and Ding, Errui},
  booktitle={Proceedings of the IEEE/CVF international conference on computer vision},
  pages={6649--6658},
  year={2021}
}

@inproceedings{artflow,
  title={Artflow: Unbiased image style transfer via reversible neural flows},
  author={An, Jie and Huang, Siyu and Song, Yibing and Dou, Dejing and Liu, Wei and Luo, Jiebo},
  booktitle={Proceedings of the IEEE/CVF conference on computer vision and pattern recognition},
  pages={862--871},
  year={2021}
}

@inproceedings{stytr2,
  title={Stytr2: Image style transfer with transformers},
  author={Deng, Yingying and Tang, Fan and Dong, Weiming and Ma, Chongyang and Pan, Xingjia and Wang, Lei and Xu, Changsheng},
  booktitle={Proceedings of the IEEE/CVF conference on computer vision and pattern recognition},
  pages={11326--11336},
  year={2022}
}

@article{swapping,
  title={Swapping autoencoder for deep image manipulation},
  author={Park, Taesung and Zhu, Jun-Yan and Wang, Oliver and Lu, Jingwan and Shechtman, Eli and Efros, Alexei and Zhang, Richard},
  journal={Advances in Neural Information Processing Systems},
  volume={33},
  pages={7198--7211},
  year={2020}
}

@inproceedings{splicevit,
  title={Splicing vit features for semantic appearance transfer},
  author={Tumanyan, Narek and Bar-Tal, Omer and Bagon, Shai and Dekel, Tali},
  booktitle={Proceedings of the IEEE/CVF conference on computer vision and pattern recognition},
  pages={10748--10757},
  year={2022}
}

@inproceedings{inst,
  title={Inversion-based style transfer with diffusion models},
  author={Zhang, Yuxin and Huang, Nisha and Tang, Fan and Huang, Haibin and Ma, Chongyang and Dong, Weiming and Xu, Changsheng},
  booktitle={Proceedings of the IEEE/CVF conference on computer vision and pattern recognition},
  pages={10146--10156},
  year={2023}
}

@inproceedings{vct,
  title={General image-to-image translation with one-shot image guidance},
  author={Cheng, Bin and Liu, Zuhao and Peng, Yunbo and Lin, Yue},
  booktitle={Proceedings of the IEEE/CVF international conference on computer vision},
  pages={22736--22746},
  year={2023}
}

@article{diffuseit,
   title={Diffusion-based Image Translation using Disentangled Style and Content Representation},
  author={Kwon, Gihyun and Ye, Jong Chul},
  journal={arXiv preprint arXiv:2209.15264},
  year={2022}
}

@inproceedings{unet,
  title={U-net: Convolutional networks for biomedical image segmentation},
  author={Ronneberger, Olaf and Fischer, Philipp and Brox, Thomas},
  booktitle={International Conference on Medical image computing and computer-assisted intervention},
  pages={234--241},
  year={2015},
  organization={Springer}
}
\end{document}